\pdfoutput=1
\documentclass{article}

\usepackage[dblblindworkshop, final]{neurips_2026}
\workshoptitle{2nd Workshop on Advances in Representation Learning for Earth Observation}

\usepackage[utf8]{inputenc}
\usepackage[T1]{fontenc}
\usepackage{hyperref}
\usepackage{url}
\usepackage{booktabs}
\usepackage{amsfonts}
\usepackage{amsmath}
\usepackage{amssymb}
\usepackage{nicefrac}
\usepackage{microtype}
\usepackage{xcolor}
\usepackage{graphicx}
\usepackage{placeins}
\usepackage{wrapfig}

\title{Detecting Weather Fronts from Satellite Observations: What Does Reanalysis Add?}

\author{%
  Ioannis Dravilas\thanks{Corresponding author: \texttt{giannisdravilas@gmail.com}} \\
  University of Amsterdam \\
  \And
  Miruna Stoicescu \\
  EUMETSAT \\
  \And
  Roope Tervo \\
  EUMETSAT \\
  \And
  Carlos Horn \\
  EUMETSAT \\
  \AND
  Oriol Hinojo Comellas \\
  EUMETSAT \\
  \And
  Sina Montazeri \\
  EUMETSAT \\
  \And
  Andrea Meraner \\
  EUMETSAT \\
  \And
  Ana Lucic \\
  University of Amsterdam \\
}

\begin{document}
\maketitle

\begin{abstract}

Weather fronts are transition regions between air masses with contrasting thermodynamic properties and are associated with mid-latitude precipitation and severe weather. We investigate machine learning (ML) methods for detecting weather fronts and quantify the contribution of reanalysis data to performance.
We compare two configurations of the same ML model for pixel-wise frontal classification over the North Atlantic and Europe: (1) using only geostationary infrared satellite observations, and (2) additionally using 28 ERA5 reanalysis fields. The satellite-only model achieves a neighbourhood critical success index at a 150\,km tolerance (CSI@150) of $0.402$, which lies within the range of agreement between analyses from different meteorological services. The satellite$+$ERA5 model achieves $0.575$, an improvement of $0.173$.
Our analysis shows that low-level winds are the least predictable from the satellite representation ($R^2 = 0.23$, compared with $0.88$ for temperature), while removing winds or moisture causes the largest decreases in detection performance.
\end{abstract}

\section{Introduction}
\label{sec:intro}

Weather fronts are narrow transition regions where air masses with contrasting thermodynamic properties meet, causing sharp changes in temperature, humidity, wind, and pressure~\citep{Hewson1998-hg, Biard2019-to}.
Detecting fronts matters for both weather forecasting and climate analysis, because a substantial fraction of extreme mid-latitude precipitation occurs near them~\citep{Catto2013-em}.
Operational forecasters identify fronts manually~\citep{Biard2019-to, Justin2025-ti} from surface observations, satellite imagery, and numerical weather prediction (NWP) output. Independent meteorological services often analyse the same synoptic situation differently, because their local information, analysis conventions, and regional priorities differ. Automated methods, from thermodynamic diagnostics on gridded fields~\citep{Hewson1998-hg, Schemm2015-xu} to deep learning models~\citep{Biard2019-to, Lagerquist2019-bg, Niebler2022-ih, Justin2025-ti}, offer objective and reproducible alternatives. Most rely on reanalysis, which combines historical observations with an NWP model to reconstruct past atmospheric states.
We study the automated pixel-wise classification of cold, warm, and occluded fronts over the North Atlantic and Europe.

Detectors that rely on reanalysis or NWP fields lag behind the latest observations. Reanalyses such as ERA5~\citep{Hersbach2020-or} become available after several days and operational NWP analyses typically after a few hours, while real-time forecasts contain no observations made after their initialisation. Geostationary satellites, in contrast, observe the atmosphere several times per hour, so a satellite-only detector can run in near real time. However, infrared radiances do not directly measure the temperature, humidity, and wind fields that characterise frontal zones. Reanalyses estimate these fields by data assimilation. This trade-off motivates our question: what does reanalysis add beyond satellite observations for frontal detection?
We answer it by comparing satellite-only and satellite$+$ERA5 detectors through cross-modal probing, variable-group ablation, and single-group addition.

\section{Related work}
\label{sec:related}

Traditional objective frontal detection applies rule-based thermodynamic diagnostics to gridded atmospheric fields~\citep{Hewson1998-hg, Schemm2015-xu}. The NWC~SAF Automatic Satellite Image Interpretation (ASII) product~\citep{NWCSAF2022-asii} detects fronts in geostationary imagery with rule-based morphological analysis, optionally adding NWP fields, and is our reference baseline.

Most deep learning approaches rely on reanalysis. \citet{Biard2019-to} and \citet{Lagerquist2019-bg} trained convolutional neural networks to detect atmospheric fronts over North America using reanalysis variables. \citet{Niebler2022-ih} used multi-level ERA5 over Europe and North America and introduced label deformation for positional uncertainty, while \citet{Justin2025-ti} developed FrontFinder~AI (UNet3+), which is deployed operationally at NOAA. None of these studies evaluates a satellite-only configuration. \citet{Qin2024-detr} combined geostationary imagery with ERA5-derived fields to detect cold fronts over East Asia, but did not consider other front types or isolate the contribution of ERA5 over satellite observations alone.

Prior work therefore evaluates a fixed input configuration~\citep{Biard2019-to, Lagerquist2019-bg, Niebler2022-ih, Justin2025-ti, Qin2024-detr}, sometimes ranking individual variables by importance~\citep{Justin2025-ti}. We instead compare satellite-only and satellite$+$ERA5 models directly and identify which ERA5 variables add information beyond the satellite representation. This matters when designing detectors capable of operating in near real time without relying on reanalysis data.

\section{Methodology}
\label{sec:method}

\subsection{Problem formulation}
Given geostationary satellite observations over the North Atlantic and Europe, optionally augmented with ERA5 reanalysis fields, we classify each grid point into one of four classes: cold front, warm front, occluded front, or no front (example in Appendix~\ref{app:example}). We use UK Met Office surface-analysis charts as ground truth because they are available every six hours over the 2019--2024 study period and their graphical conventions allow more reliable automated digitisation than the charts of some other services. We use data from four additional services as independent baselines (Appendix~\ref{app:baselines}).

We train two models with a Swin-UNet~\citep{Liu2021-ng, Cao2023-swinunet} inspired by Aurora~\citep{Bodnar2025-ze} (Appendix~\ref{app:arch}) that differ only in their inputs.
These are the single-service variants of the SATFRONT detector~\citep{Dravilas2026-satfront}, trained only on UK Met Office labels. The main SATFRONT model instead uses labels from five services.
The satellite-only model (\textbf{OBS}) uses 15 satellite-derived channels: eight infrared channels (3.8--13.3\,$\mu$m) from the GEO-Ring geostationary data record~\citep{EUMETSAT2024-georing} at approximately 5\,km resolution on the equal-area HEALPix grid, and seven deterministic features derived from the brightness temperatures following the ASII scheme~\citep{NWCSAF2022-asii}. The satellite$+$ERA5 model (\textbf{HYB}) adds 28 ERA5 fields~\citep{Hersbach2020-or}: six atmospheric variables at four pressure levels (1000--700\,hPa) and four additional surface variables. For analysis, we group the ERA5 variables into winds (10 fields), moisture (8), temperature (5), and mass (5). Appendix~\ref{app:arch} lists all input channels. Because ERA5 becomes available only after several days, we use HYB as a diagnostic upper bound on what reanalysis adds to this detector, not as an intended operational model. We train on 2019--2022, validate on 2023, and test on 2024.

\subsection{Experimental setup}
We first compare the satellite-only and satellite$+$ERA5 models, then use three analyses to identify which ERA5 variables explain the performance difference: (i) cross-modal probing tests which ERA5 fields the satellite-only encoder can already predict, (ii) variable-group ablation measures the effect of removing each ERA5 group from HYB, and (iii) single-group addition measures the gain from adding each ERA5 group alone to OBS.

\begin{wrapfigure}{r}{0.40\linewidth}
  \centering
  \includegraphics[width=\linewidth]{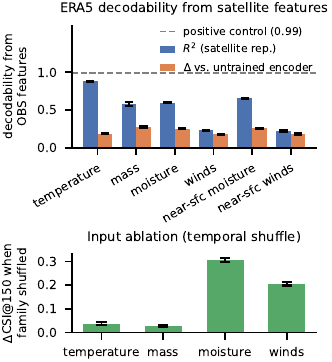}
  \caption{Analysis of ERA5 variable groups. \emph{Top:} linear predictability ($R^2$) from the frozen satellite-only encoder (best stage) and gain over a randomly initialised encoder, with 95\% CIs. Near-sfc: 1000/925\,hPa and 10\,m fields. Dashed: positive control. \emph{Bottom:} drop in CSI@150 when a group is replaced by values from another date.}
  \label{fig:analysis}
\end{wrapfigure}

\paragraph{Cross-modal probing.} We freeze the OBS encoder and extract features from each of its four hierarchical stages. Successive stages reduce spatial resolution and increase channel dimension. At each stage, we fit per-pixel ridge regression to predict every ERA5 field from the encoder features. Because variables may be best resolved at different scales, we take each field's highest coefficient of determination ($R^2$) on the 2024 test set across the four stages and average it within each variable group. To separate what training adds from what the inputs or architecture provide, we repeat the procedure with the raw satellite input and a randomly initialised encoder (Appendix~\ref{app:probe}).

\paragraph{Variable-group ablation.} For each variable group, we replace its HYB input channels with values from a randomly selected date and measure the change in CSI@150 relative to the unmodified HYB model. The replacement keeps each field's spatial distribution but breaks its temporal alignment with the frontal labels.

\paragraph{Single-group addition.} We train three additional models, each combining the satellite inputs with one ERA5 variable group, using the same architecture and training procedure as OBS and HYB. We add temperature and mass as one group because, by the hypsometric equation, the geopotential thickness between two pressure levels is set by the mean virtual temperature of the layer~\citep{Holton2013-dynmet}. This grouping also limits the number of new training runs. We measure each model's paired CSI@150 gain over OBS.

\paragraph{Evaluation.} We report the critical success index at a 150\,km tolerance (CSI@150), following \citet{Justin2025-ti}. A ground-truth front pixel counts as detected if a predicted pixel of the same class lies within 150\,km, and a predicted pixel counts as a false positive if no ground-truth pixel of the same class lies within 150\,km. We compute CSI per class and average over the three classes. We estimate 95\% confidence intervals for the difference $\Delta$ between two models (e.g., HYB $-$ OBS) with a paired moving-block bootstrap (Appendix~\ref{app:probe}) and report $P(\Delta \le 0)$, the fraction of bootstrap samples with zero or negative improvement. Every paired comparison with OBS, HYB, or the random encoder in Section~\ref{sec:results} has $P(\Delta \le 0) < 10^{-3}$. All results use a single training seed. We compare the models with the operational ASII product~\citep{NWCSAF2022-asii} and with the agreement between analyses from different meteorological services, measured against the same ground truth (Appendix~\ref{app:baselines}).

\FloatBarrier
\section{Results and discussion}
\label{sec:results}

\paragraph{Comparison of satellite-only and satellite$+$ERA5 models.}
Table~\ref{tab:gap} compares OBS and HYB with two reference baselines (Appendix~\ref{app:baselines}). OBS reaches CSI@150 $= 0.402$, approximately three times the ASII score ($0.143$). The OBS score also lies within the range of agreement between analyses from different meteorological services ($0.277$ to $0.501$), and Met~\'Eireann agrees most closely with the UK Met Office. These comparisons provide context, not an upper bound. The services differ in local information, conventions, and geographic priorities, whereas we train and score OBS on the UK Met Office convention. HYB reaches $0.575$ ($\Delta = +0.173$, 95\% CI $[0.163, 0.183]$), a $43\%$ improvement over OBS. The gain is similar across front types ($+0.164$, $+0.174$, and $+0.180$ for cold, warm, and occluded fronts) and comes mainly from higher probability of detection ($0.505$ to $0.718$).

\begin{table}[t]
  \centering
  \caption{CSI@150 on the 2024 test year (Met Office ground truth). Overall is the mean over the three front types; brackets give 95\% CIs. All per-class improvements exclude zero ($P(\Delta \le 0) < 10^{-3}$).}
  \label{tab:gap}
  \small
  \begin{tabular}{lcccc}
    \toprule
    & Overall & Cold & Warm & Occluded \\
    \midrule
    Operational ASII product (satellite$+$NWP) & 0.143 & 0.156 & 0.109 & 0.163 \\
    Closest human analysis (Met~\'Eireann) & 0.501 & 0.543 & 0.454 & 0.508 \\
    \midrule
    OBS (satellite-only) & 0.402 \tiny{[0.389, 0.414]} & 0.450 & 0.359 & 0.398 \\
    HYB (satellite$+$ERA5) & 0.575 \tiny{[0.563, 0.585]} & 0.615 & 0.532 & 0.578 \\
    ERA5 improvement $\Delta$ & +0.173 \tiny{[0.163, 0.183]} & +0.164 & +0.174 & +0.180 \\
    \bottomrule
  \end{tabular}
\end{table}

\paragraph{Contribution of ERA5 variable groups.}
In the probe (Figure~\ref{fig:analysis}, top), temperature is the ERA5 group that the frozen OBS encoder predicts best ($R^2 = 0.88$), followed by moisture ($0.60$) and mass ($0.59$), while winds are the least predictable ($R^2 = 0.23$). High temperature predictability is consistent with the direct thermal information in infrared radiances, whereas the encoder can infer wind only indirectly from cloud and moisture patterns~\citep{Velden2005-winds}. A randomly initialised encoder already shows the same ranking ($R^2 = 0.69$ for temperature and $0.05$ for winds). Training improves all groups by $+0.18$ to $+0.28$ but does not close the gap for winds.

In the ablation (Figure~\ref{fig:analysis}, bottom), replacing winds or moisture lowers CSI@150 by $0.21$ and $0.31$, whereas replacing temperature or mass causes much smaller decreases ($0.04$ and $0.03$). Replacing one group also breaks its correlations with the other groups, such as the thermal-wind relationship between winds and temperature. We therefore compare the wind-and-moisture block with the temperature-and-mass block and do not rank the groups within each block.

In the single-group addition experiments (Appendix~\ref{app:fusion}), adding each group alone to OBS yields similar gains: $+0.118$ for winds, $+0.138$ for moisture, and $+0.132$ for temperature and mass. Temperature and mass have little effect when ablated but provide comparable gains when added alone, consistent with strong correlations among the ERA5 groups. Winds are the clearest case in which low predictability from the satellite encoder coincides with a large ablation effect.

\begin{wrapfigure}{r}{0.29\linewidth}
  \centering
  \includegraphics[width=\linewidth]{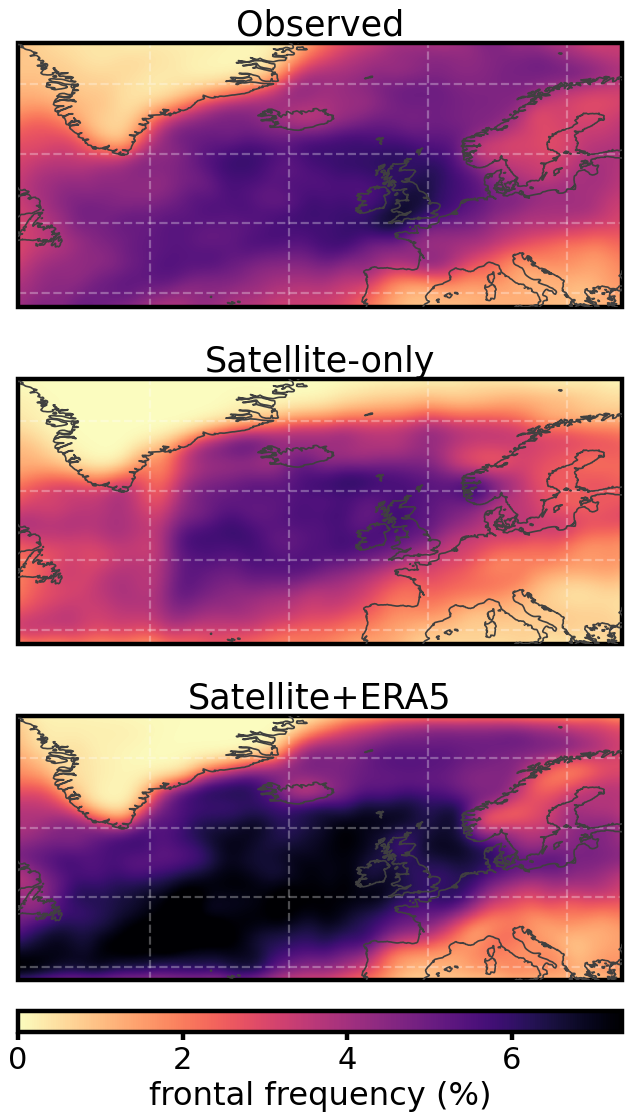}
  \caption{Annual frontal frequency (2024, all front types, smoothed at ${\sim}150$\,km for display).}
  \label{fig:clim}
\end{wrapfigure}

\paragraph{Effect on detection characteristics.}
The improvement from ERA5 grows with the tolerance, from $+0.116$ at 50\,km to $+0.173$ at 150\,km, while the gain in probability of detection stays nearly constant (Appendix~\ref{app:radius}). This pattern is consistent with HYB detecting fronts that OBS misses. OBS also partly reproduces the annual Met Office frontal-frequency map (Pearson correlation $0.37$), and ERA5 raises this correlation to $0.53$ (Figure~\ref{fig:clim}). Among front types, occluded fronts show the largest improvement in front placement in individual analyses ($+0.180$), but ERA5 reduces the mean-squared error (MSE) of their annual frequency map by only 5.5\%. Cold and warm fronts improve on both measures, with MSE reductions of 47\% and 39\% for the single-service models used here. With the five-service models of \citet{Dravilas2026-satfront}, the cold-front MSE instead rises by 3.5\%.

\paragraph{Limitations and conclusion.}
Our results show that satellite observations alone provide substantial information for frontal detection, while adding ERA5 raises CSI@150 from $0.402$ to $0.575$. Our analyses help explain this gap. The probe and the ablation agree most clearly on low-level winds. Winds are the ERA5 variables least predictable from the satellite representation and, together with moisture, those whose replacement most reduces performance. For occluded fronts, ERA5 strongly improves detection on individual charts (CSI@150) but barely reduces the error of the annual frequency map, consistent with reanalysis providing instantaneous dynamical information that infrared imagery does not directly measure.

These conclusions remain subject to several limitations. Linear probing measures which information the representation encodes, not what the decoder actually uses~\citep{Ravichander2021-probing}, and our ablations disrupt correlations among ERA5 variables. We also evaluate one domain, architecture, and training seed. Nevertheless, cross-modal probing of a frozen unimodal encoder can estimate how much of a candidate modality the existing representation already predicts, before any multi-modal model is trained. The same approach could be applied to other combinations of observations and variables in Earth observation.

\FloatBarrier

\begin{ack}
We thank the Finnish Meteorological Institute (FMI) for providing its frontal analyses. We also acknowledge the German Weather Service (DWD), the Royal Netherlands Meteorological Institute (KNMI), the Irish Meteorological Service (Met~\'Eireann), and the UK's national meteorological service (UK Met Office), whose surface-analysis charts we used in this work. We acknowledge EUMETSAT, Destination Earth, and the European Weather Cloud for providing compute and storage resources. The authors declare no competing interests.
\end{ack}

\bibliographystyle{plainnat}
\bibliography{references}

\appendix

\section{Model architecture and input channels}
\label{app:arch}

Both configurations use a Swin-UNet encoder--decoder segmentation network~\citep{Liu2021-ng, Cao2023-swinunet}, adapted to two dimensions from the three-dimensional Swin U-Net processor in Aurora~\citep{Bodnar2025-ze}. The encoder uses a convolutional patch-embedding stem followed by four stages. After the first stage, each subsequent stage halves the spatial resolution and doubles the channel width. Each stage applies windowed self-attention with alternating shifted windows, allowing information to propagate across windows. A symmetric decoder with skip connections restores the original grid resolution. Unlike Aurora, our model does not use a three-dimensional pressure-level axis, Perceiver encoding, or foundation-model pre-training. \citet{Dravilas2026-satfront} give a more detailed description of the architecture. The single-service variants used here have one output head, whereas the five-service model SATFRONT has one per service.

The satellite input contains eight infrared brightness temperatures at approximate central wavelengths of 3.8, 6.2, 7.3, 8.6, 9.7, 11.0, 12.0, and 13.3\,$\mu$m, together with seven auxiliary features. Following ASII~\citep{NWCSAF2022-asii}, we compute these features from the brightness temperatures of the 6.2, 11.0, and 12.0\,$\mu$m channels. The first two are differences between channels and are sensitive to upper-atmospheric moisture and cloud properties. The remaining five are spatial filters of the 11.0\,$\mu$m channel that highlight local anomalies, cloud texture, cloud edges, curved cloud structures, and large-scale cloud bands. All seven features use satellite data only. The 28 ERA5 fields comprise geopotential height, specific humidity, relative humidity, temperature, and horizontal wind components ($u$, $v$) at four pressure levels (1000, 925, 850, 700\,hPa), plus 2-metre temperature, mean sea level pressure, and 10-metre wind components ($u_{10}$, $v_{10}$). OBS uses the 15 satellite channels, while HYB uses all 43 channels (15 satellite plus 28 ERA5 fields). We map the satellite observations from the HEALPix grid to a regular $0.05^\circ$ latitude--longitude grid using nearest-neighbour interpolation and bilinearly interpolate the ERA5 fields to the same grid.

\section{Reference baselines}
\label{app:baselines}

The two reference rows in Table~\ref{tab:gap} provide external context. We score the operational NWC~SAF ASII product~\citep{NWCSAF2022-asii}, a rule-based satellite$+$NWP front detection product (the ASII-NWP variant), against the same 2024 Met Office charts (1453 of the 1462 test timestamps) using the same CSI@150 metric as for the learned models. The Met~\'Eireann row shows the agreement between analyses from different meteorological services. To measure this agreement, we score the 2024 surface-analysis charts of four other services in place of the model predictions, against the charts of the UK's national meteorological service (UK Met Office) used as ground truth throughout. The four services are the Finnish Meteorological Institute (FMI), the German Weather Service (DWD), the Royal Netherlands Meteorological Institute (KNMI), and the Irish Meteorological Service (Met~\'Eireann). For each service, we evaluate only the overlap with the Met Office domain and timestamps for which both services provide an analysis (359 for Met~\'Eireann, 366 for DWD, 893 for FMI, and 1460 for KNMI), so these scores do not use exactly the same samples as the models. Agreement ranges from $0.277$ (FMI) to $0.501$ (Met~\'Eireann). This variation reflects not only differences in frontal positioning but also differences in analysis conventions, available local information, and geographic focus, as each service primarily analyses its own region of interest. The range therefore provides context for model performance but does not represent a ceiling on achievable detection skill. No directly comparable machine learning baseline exists because no prior detector has been trained and evaluated on the same geographic domain, time period, and labelling convention.

\section{Representation analysis details}
\label{app:probe}

We freeze the OBS encoder and extract features from each of its four stages. At each stage, we fit a per-pixel ridge-regression model to predict each of the 28 ERA5 fields from the encoder features, using 500 training-year timestamps sampled at regular intervals. We pool each ERA5 target to the corresponding feature resolution using adaptive average pooling and subsample pixels by a factor of four during fitting. We standardise all features before regression. We select the ridge penalty from $\{0.1, 1, 10, 100\}$ using a 20\% held-out split of the training subset. We evaluate $R^2$ on the 2024 test year, take the maximum for each field across the four encoder stages so that each variable is evaluated at its most favourable resolution, and then average within each variable group. We compute all 95\% confidence intervals in this paper using a paired moving-block bootstrap over the 1462 test timestamps (block length~20, $B = 1000$ resamples). A value of $P(\Delta \le 0) < 10^{-3}$ means that no resample gave $\Delta \le 0$. All results use a single training seed.

We repeat the same procedure with two controls. The first replaces the encoder features with the raw satellite input, pooled to each feature resolution. The second uses features from two independently randomly initialised encoders with the same architecture and averages $R^2$ across the two. These controls distinguish the contribution of the trained representation from that of the inputs or architecture alone. As a verification, predicting two of the model's own input channels from its encoder features gives $R^2 \approx 0.99$. Table~\ref{tab:probe} reports the results by variable group.

\begin{table}[h]
  \centering
  \caption{Predictability of ERA5 variables from the frozen OBS encoder, by variable group (2024 test year). $R^2$ is the maximum over encoder stages. The $\Delta$ columns show the improvement over the randomly initialised encoder and the raw satellite input. All improvements exclude zero ($P(\Delta \le 0) < 10^{-3}$). 95\% confidence intervals for $R^2$ and the improvement over the random encoder are shown in Figure~\ref{fig:analysis}.}
  \label{tab:probe}
  \small
  \begin{tabular}{lccc}
    \toprule
    Variable group & $R^2$ (max over stages) & $\Delta$ vs.\ random init.
                   & $\Delta$ vs.\ raw input \\
    \midrule
    Temperature & 0.883 & $+0.188$ & $+0.177$ \\
    Moisture    & 0.600 & $+0.254$ & $+0.244$ \\
    Mass        & 0.585 & $+0.282$ & $+0.234$ \\
    Winds       & 0.232 & $+0.181$ & $+0.189$ \\
    \bottomrule
  \end{tabular}
\end{table}

\FloatBarrier
\section{Single-group addition experiments}
\label{app:fusion}

Table~\ref{tab:fusion} reports CSI@150 for the satellite-only model augmented with one ERA5 variable group, together with the paired improvement over OBS.

\begin{table}[h]
  \centering
  \caption{Single-group addition experiments. CSI@150 of the satellite-only model augmented with one ERA5 variable group, and the paired improvement over OBS (2024 test year, 95\% paired block-bootstrap confidence intervals). All three improvements have $P(\Delta \le 0) < 10^{-3}$.}
  \label{tab:fusion}
  \small
  \begin{tabular}{lccc}
    \toprule
    & OBS + winds & OBS + moisture & OBS + temp.\ \& mass \\
    \midrule
    CSI@150 & 0.520 \tiny{[0.506, 0.530]} & 0.540 \tiny{[0.527, 0.550]}
            & 0.534 \tiny{[0.521, 0.545]} \\
    $\Delta$ over OBS & +0.118 \tiny{[0.108, 0.126]} & +0.138 \tiny{[0.129, 0.147]}
                      & +0.132 \tiny{[0.122, 0.142]} \\
    \bottomrule
  \end{tabular}
\end{table}

\begin{figure}[t]
  \centering
  \includegraphics[width=\linewidth]{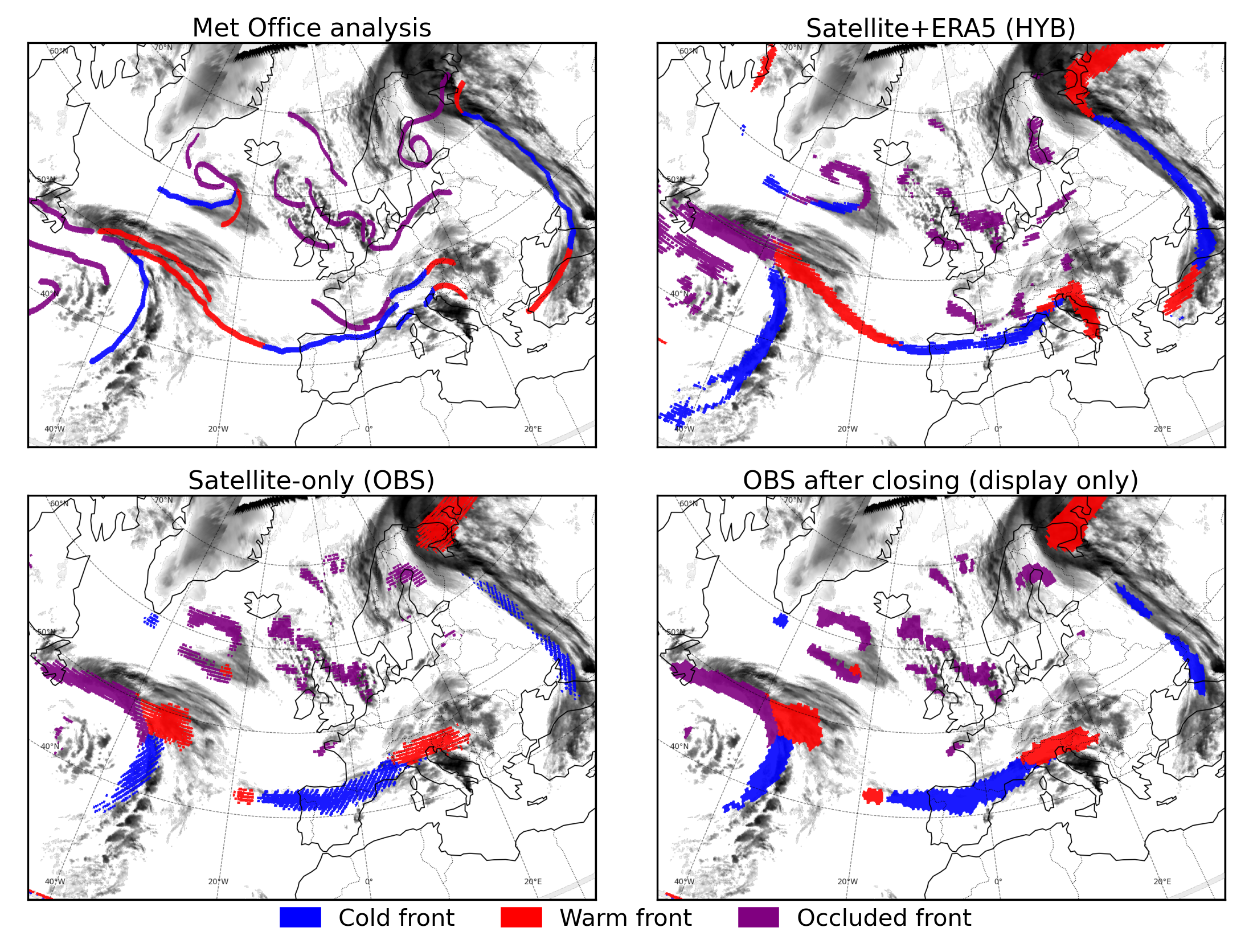}
  \caption{Example on one 2024 test timestamp (22 November, 00:00~UTC). \emph{Top left:} Met Office analysis. \emph{Top right:} satellite$+$ERA5 model (HYB). \emph{Bottom left:} satellite-only model (OBS). \emph{Bottom right:} the same OBS prediction after a morphological closing that bridges small gaps and removal of small regions, applied for visualisation only. All reported metrics use the unprocessed predictions. Cold, warm, and occluded fronts are shown in blue, red, and purple over the infrared background.}
  \label{fig:example}
\end{figure}

\section{ERA5 improvement across tolerances}
\label{app:radius}

Table~\ref{tab:radius} reports the paired improvement in CSI from adding ERA5 (HYB $-$ OBS) at each tolerance. The gain is smallest at 50\,km ($+0.116$), peaks at 150\,km ($+0.173$), and decreases slightly to $+0.165$ at 250\,km. However, because both models achieve higher CSI at larger tolerances, the absolute gain alone cannot distinguish additional detections from improved positioning. The relative gain also decreases with tolerance, from $48\%$ at 50\,km to $43\%$ at 150\,km and $32\%$ at 250\,km.
CSI combines two components, the probability of detection (the fraction of observed front pixels matched by predictions) and the success ratio (the fraction of predicted front pixels matching observed fronts). The gain in probability of detection remains nearly constant, between $+0.20$ and $+0.21$, across all tolerances from 50 to 250\,km. If ERA5 primarily improved the positioning of fronts already detected by OBS, this gain would diminish at larger tolerances as displaced OBS predictions begin to match the observations. Instead, the persistent gain is consistent with HYB detecting fronts missed entirely by OBS, which count as additional detections regardless of tolerance. In contrast, the success-ratio gain decreases from $+0.16$ at 50\,km to $+0.05$ at 250\,km, indicating that HYB also places its predicted fronts closer to the analysed fronts.
\begin{table}[h]
  \centering
  \caption{Paired ERA5 improvement in CSI (HYB $-$ OBS) by tolerance, 2024 test year, with 95\% paired block-bootstrap confidence intervals. All exclude zero ($P(\Delta \le 0) < 10^{-3}$).}
  \label{tab:radius}
  \small
  \begin{tabular}{lccccc}
    \toprule
    Tolerance (km) & 50 & 100 & 150 & 200 & 250 \\
    \midrule
    $\Delta$CSI & $+0.116$ & $+0.160$ & $+0.173$ & $+0.172$ & $+0.165$ \\
    95\% CI & \tiny{[0.109, 0.124]} & \tiny{[0.151, 0.169]} & \tiny{[0.163, 0.183]}
            & \tiny{[0.162, 0.183]} & \tiny{[0.155, 0.177]} \\
    \bottomrule
  \end{tabular}
\end{table}

\section{Qualitative example}
\label{app:example}

The main-text evaluation reports quantitative results averaged over the full 2024 test year. Figure~\ref{fig:example} instead illustrates OBS and HYB predictions for a single test timestamp. The top-left panel shows the Met Office analysis. The top-right panel shows HYB, which reproduces the frontal structure more completely and follows the Met Office analysis more closely. The bottom-left panel shows OBS, which detects the main systems but breaks many fronts into short, scattered segments and misses others. These fragments make the frontal structure hard to read, so the bottom-right panel shows the same OBS prediction after a morphological closing that bridges gaps of up to about 12 pixels, followed by removal of connected regions smaller than 120 pixels. We apply this step only to OBS, because the HYB predictions already form continuous fronts. This post-processing is for visualisation only (as in an operational display) and does not affect any reported metric, which we always compute on the unprocessed predictions of both models.

\end{document}